\documentclass[journal]{IEEEtran}
\usepackage{amsmath,amsfonts}
\usepackage{algorithmic}
\usepackage{algorithm}
\usepackage{array}
\usepackage[caption=false,font=normalsize,labelfont=sf,textfont=sf]{subfig}
\usepackage{textcomp}
\usepackage{stfloats}
\usepackage{url}
\usepackage{verbatim}
\usepackage{graphicx}
\usepackage{cite}
\usepackage{multirow}

\begin{document}

\title{Optical Tactile Sensing for Aerial Multi-Contact Interaction: Design, Integration, and Evaluation}

\author{Emanuele Aucone, Carmelo Sferrazza,~\IEEEmembership{Member,~IEEE}, Manuel Gregor, \\ Raffaello D'Andrea,~\IEEEmembership{Fellow,~IEEE} and Stefano Mintchev,~\IEEEmembership{Member,~IEEE}
\thanks{Emanuele Aucone, Manuel Gregor and Stefano Mintchev are with the Environmental Robotics Laboratory, Dept. of Environmental Systems Science, ETH Z\"urich, 8092 Z\"urich, Switzerland and with the Swiss Federal Institute for Forest, Snow and Landscape Research (WSL), 8903 Birmensdorf, Switzerland (e-mail: \textit{eaucone@ethz.ch}). 
Carmelo Sferrazza is with the Robot Learning Lab, UC Berkeley, CA 94704 Berkeley, USA. 
Raffaello D'Andrea is with the Institute for Dynamic Systems and Control, Dept. of Mechanical and Process Engineering, ETH Z\"urich, 8092 Z\"urich, Switzerland. 
This work was supported by the Swiss National Science Foundation through the Eccellenza Grant number 186865.}
}

\markboth{IEEE Transactions on Robotics, ~Vol.~YY, No.~X, MM~202Z}%
{Aucone \MakeLowercase{\textit{et al.}}: Optical Tactile Sensing for Aerial Multi-Contact Interaction}


\maketitle


\begin{abstract}
Distributed tactile sensing for multi-force detection is crucial for various aerial robot interaction tasks. However, current contact sensing solutions on drones only exploit single end-effector sensors and cannot provide distributed multi-contact sensing. Designed to be easily mounted at the bottom of a drone, we propose an optical tactile sensor that features a large and curved soft sensing surface, a hollow structure and a new illumination system. Even when spaced only 2 cm apart, multiple contacts can be detected simultaneously using our software pipeline, which provides real-world quantities of 3D contact locations (mm) and 3D force vectors (N), with an accuracy of 1.5 mm and 0.17 N respectively. We demonstrate the sensor's applicability and reliability onboard and in real-time with two demos related to i) the estimation of the compliance of different perches and subsequent re-alignment and landing on the stiffer one, and ii) the mapping of sparse obstacles. The implementation of our distributed tactile sensor represents a significant step towards attaining the full potential of drones as versatile robots capable of interacting with and navigating within complex environments.

\end{abstract}

\begin{IEEEkeywords}
System-level integration of tactile sensors in robots, Aerial robots: Applications, Tactile sensors design, Physical interactions by touching.
\end{IEEEkeywords}


\section{Introduction}
\label{sec:introduction}

\IEEEPARstart{D}{rones} have become an increasingly popular tool in various fields, including infrastructure inspection and maintenance, agriculture, search and rescue, environmental monitoring, education and more \cite{drones_floreano_wood, search_and_rescue, future_farms, drone_wildlife}. Safely flying in open spaces is currently mastered through GPS or vision-based sensors \cite{agile_flight_wild, swarm_robots_wild}. Yet, in cluttered environments or in visually-degraded settings (e.g., smoke, fog, low light), these solutions have limited ability to interpret and perceive information about the surroundings \cite{occluded_branch}. The high chance of collision with obstacles prompts the exploration of touch-based perception alternatives for effective navigation \cite{briod,contact_nav_alexis}. Moreover, drones are often required to physically interact with structures such as tubes and pipes, during industrial inspection and maintenance \cite{omav_inspection,maintanence}, as well as branches, for monitoring and sampling in natural environments \cite{sensor_installation, gripper_sensor, aucone}. To this end, the sense of touch is essential, as it provides robots with the ability to measure external interactions with the environment \cite{tactile_sensing}. 

The integration of rich tactile sensing in the from of large-area electronic skin patches or high-resolution touch sensors has been steadily growing in robotic hands, arms and humanoids for manipulation tasks and interaction with humans \cite{tactile_manupulation_review, tactile_skin_review, touch}. Tactile sensors also found increasing uses in terrestrial robots for locomotion, navigation and terrain classification \cite{whisker_robot, haptic-inspection-soil, vangen2023terrain}. Despite these advancements, the sense of touch remains surprisingly underutilized in drones. The integration of rich tactile sensing in drones would allow the detection of multiple contacts around the drone's body, which could be beneficial to a variety of applications, e.g. to apply multiple forces on surfaces (interaction/control task), to track and move objects along a trajectory (non-prehensile manipulation), to estimate the location of the surrounding obstacles and have a comprehensive perception of the environment (mapping), to estimate the compliance of the surroundings and find a safe path to traverse obstacles (navigation). This motivates our work, which presents the development of a large-area tactile sensor for use onboard drones to simultaneously measure multiple contact locations and forces when interacting with the surrounding environment. We believe that our work paves the road for future designs, integration and exploitation of optical tactile sensors for a wide range of drone applications.

\vspace{-0.2cm}
\subsection{Related Work}

The exploitation of force feedback in drones has brought to significant achievements in the field of Aerial Physical Interaction (APhI) \cite{aerial_robotic_manipulators_review_Ollero}. By exploiting a single multi-axis load-cell to measure the external wrench (forces and torques) exerted by the environment during contact, researchers have demonstrated complex tasks such as contact-based inspection \cite{inspection, omav_inspection}, aerial manipulation \cite{Fumagalli2014,Bartelds2016,Nava2020}, or pushing rolling carts \cite{dob_moving_structure,tank-based_control_uncertain_dynamic_env}. In these studies, the load cell is typically mounted at end-effector (EE) where the interaction is localized. In other works, the load cell is connected to a protective cage to measure a net wrench across the entire surface of the cage. For instance, in \cite{aucone}, a force sensing cage enables a drone to land on flexible tree branches, while in \cite{traversal}, the authors exploit it to dampen the oscillations of compliant obstacles while traversing them. However, load cells are usually expensive and often require additional instrumentation for data conversion. More importantly, a single load cell only allow the measurement of a single wrench. In case of multiple interactions, this sensor provides only a low-dimensional, aggregated information consisting of the sum of multiple interaction forces. In multi-contact interaction scenarios drones need to independently measure contact locations and interaction forces occurring over a large surface of their body, thereby requiring distributed haptic sensing with high spatial resolution.

In the scope of multiple contacts sensing, a ring-shaped sensorized structure mounted around the drone is introduced for human-robot interaction in \cite{Rajappa2017}. This ring incorporates eight push-button switches to roughly estimate the position of point of contact, but it cannot provide an estimate of the interaction force. In \cite{contact_nav_briod}, eight miniature single axis force sensors are integrated on a protective cage around the drone. This approach provides measures of individual impact forces and a coarse estimate of the contact direction, useful for contact-based navigation. For the same purpose, a bio-inspired system based on artificial whiskers is proposed in \cite{whisker_nav_alexis}. Four haptic sensors are exploited to estimate both the location of contacts and the interaction forces around the drone's body. These solutions, however, exploit minimalist sensors that allow to measure only a single-axis force. Furthermore, a high number of sensors - an array - is required in order to provide distributed sensing around the drone's body with adequate resolution, which increases the weight and the complexity of the system both in terms of components integration and computational burden.

Optical (or vision-based) tactile sensors offer a viable solution to mitigate these flaws, with the promise of high-resolution tactile sensing without relying on complex sensor arrays \cite{tactile_sensors, optical_tactile_review}. Such sensors generally consist of a contact area made of flexible material, an internal light source, and a detector, usually in the form of a camera. The latter is used to detect deformations of the contact area. Tactile information including contact location, contact force distribution or object texture can be reconstructed from images via analytical \cite{Zhang2019} or data-driven learning techniques \cite{Sferrazza2019}. The minimal number of electronic components and reduced wiring make them basically immune to electromagnetic interference and potentially lighter than sensor arrays, a feature particularly appealing in aerial robotics. Further, since the sensing elements are not integrated into the skin, they are potentially more suitable for tasks where robots physically interact with unstructured obstacles.

In this regard, the GelSight \cite{Yuan2017, vitac} sensor can measure the surface texture of objects, by tracking light intensity of a reflective surface. Recently, researchers have demonstrated the application of such sensors on the EE of a drone for wall texture detection \cite{tactile_UAV}. Other technologies include GelSlim \cite{Donlon2018} and Digit \cite{Lambeta2020}, where the design has been further miniaturized for integration on robotic fingertips. The TacTip family of sensors \cite{tactip2018}, instead, can adapt to different shapes by utilizing change in the positions of markers to provide the approximate edge of a contacted surface, for helping robots to localize and follow the contours. Exploiting markers embedded in the flexible membrane has been further validated in \cite{Lin2019}, where a reflective membrane is used in conjunction with a set of yellow and magenta colored markers, in \cite{Sferrazza2019}, where fluorescent markers are randomly scattered within a transparent material below the sensing surface to achieve very high spatial and force resolution, and in \cite{Abad2020,Kim2022}, where instead regularly spaced painted or printed ultraviolet (UV) markers are exploited. Most of the aforementioned sensors, however, have been limited to relatively small areas with simple geometries, such as fingertips or grippers, useful to estimate only the distribution of the total applied force, and their scalability to large and curved areas has not been reported. In this regard, TacLINK \cite{taclink1, Luu2023} presents an example of large optical tactile sensor using markers on the outer cylindrical surface and two cameras at the ends of the cylinder. The sensor is designed to provide tactile sensing across the whole length of a robot arm segment. Despite validating detection of multiple contacts over the large surface, the method decreases its accuracy in force estimation for deformations larger than the surface of a fingertip. In \cite{tactile_bed}, the authors introduced a bed-size tactile sensor to monitor and classify sleep position. The accuracy achieved despite the dimensions of the sensor strongly motivates us to pursue the development of a large-scale tactile sensor for aerial robots. 

\begin{figure}
    \centering
    \includegraphics[width=\linewidth]{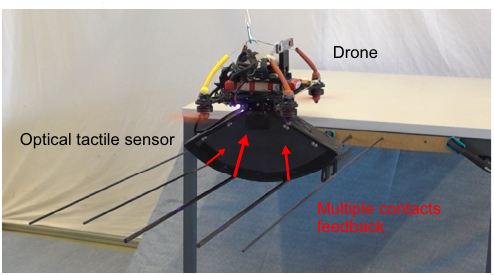}
    \caption{Optical tactile sensing for drones. A quadrotor embodying the proposed optical tactile sensor to detect multiple contacts and interaction forces during physical interaction with the environment.}
    \label{fig:main}
    \vspace{-0.3cm}
\end{figure}

\vspace{-0.2cm}
\subsection{Contribution}

In this study, we introduce a distributed tactile sensor designed for multi-contact interaction tasks with aerial robots (Fig. \ref{fig:main}). We propose novel hardware (HW) and software (SW) solutions for vision-based tactile sensing, custom-tailored for application in aerial robots. Subsequently, we showcase the utilization of this sensor in tasks involving multi-contact interaction and navigation.

From a design perspective, to enable the simultaneous detection of multiple contacts, we propose a curved, large-scale sensor with a sensing surface measuring 32 cm by 4 cm. Positioned at the bottom of the drone, the sensor serves both as a protective structure and an end-effector with distributed and high-resolution tactile sensing. The sensor has a hollow structure to minimize weight. Additionally, we introduce a novel UV-based illumination technique to provide uniform illumination of the large sensing area.

From a computational perspective, we introduce the first application of the natural Helmholtz Hodge Decomposition (nHHD) \cite{Bhatia2014} in a multi-contact scenario. Multiple contact points are characterized by their absolute location and 3D contact force vector, comprising both shear and pressure components. 

We validate the integration and real-time usage of the sensor onboard our custom drone to extend its sensing capabilities. Further, two demos are performed to demonstrate how the sensor feedback can be used for decision-making (estimation of the compliance of two perches and re-alignment and landing on the stiffer one), as well as perception of the surrounding environment (mapping of sparse obstacles).


\section{System Design}
\label{sec:design}

Optical tactile sensing provides distributed measurements over the area covered by the flexible skin. In principle, there are no assumptions made on the size and shape of the skin, as long as this is covered by the field of view of the optical unit. However, in practice, the application on aerial systems introduces the need for non-trivial design adjustments, which we addressed in this work. This chapter reports the requirements, design, and fabrication of our sensor.



\vspace{-0.2cm}
\subsection{Design rationale}

The mechanical design of the sensor draws inspiration from \cite{Sferrazza2019}. To optimize its functionality for drones, we identify four crucial modifications: i) the size of the sensing area has to be increased; ii) the shape of the sensing area is no longer flat but has a shallow curve; iii) the inner side of the sensor is hollow to make the system more lightweight; iv) the internal illumination is based on UV LEDs and placed outside the silicone. The rationale for these modifications is described below.

\begin{figure}
    \centering
    \includegraphics[width=\linewidth]{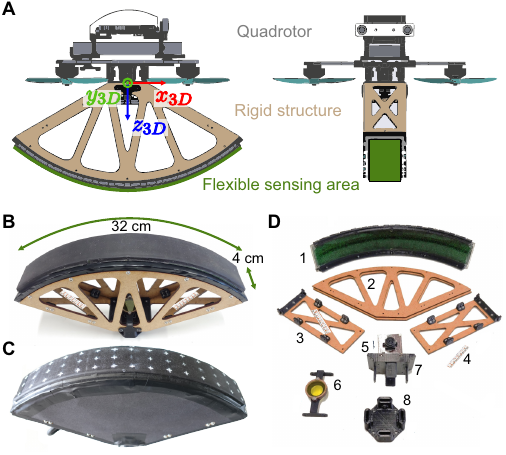}
    \caption{Mechanical design and main components of the optical tactile sensor. (A) Developed prototype of optical tactile sensor shaped as an arc of a circular ring, integrated beneath a quadrotor (side and front view). The sensing area (green) is on the outer side of the sensor. The sensor coordinate frame is highlighted. (B) Assembled sensor with internal view exposed. (C) Fully assembled sensor with paper shield on the sides; sensing area marked in different contact locations. (D) Exploded view of all the components: 1) Sensing area (acrylic substrate and silicone material) with two connectors glued on; 2) lateral panels and 3) front panels for narrow side with M2 brackets and the other connectors screwed on; 4) UV LEDs strips; 5) USB camera; 6) color filter with holder; 7) camera holder and sensor base; 8) connector to attach the sensor to the drone. }
    \label{fig:design}
    \vspace{-0.2cm}
\end{figure}

In detail, we propose a large-scale sensor shaped as an arc of a circular ring that partially covers and protects the drone (Fig. \ref{fig:design}A). 
The main structure of the sensor consists of a rigid frame and presents a mounting system that facilitates the integration on the drone's frame.
The sensing area is soft and placed on the outer side of the sensor, lying on a transparent rigid plate that follows the contour of the arc. In contrast to most of previous work that has targeted robotic fingertips, we increase the size of the sensing surface to 32 cm by 4 cm for simultaneous interaction with multiple distant obstacles (Fig. \ref{fig:design}B).
The choice of a curved surface, rather than a flat one, is motivated by previous works that have demonstrated how the usage of streamlined, sensorized cages - hemispherical \cite{aucone} or ring-shaped \cite{traversal} - can be beneficial for drones to safely interact with obstacles. 
To keep the sensor lightweight, its inner part is hollow, differently from the design in \cite{Sferrazza2019}. This means that the space between the camera and the sensing area is completely empty so that the overall weight is significantly reduced (Fig. \ref{fig:design}B) - fixing the size and the rigid components of the structure, the hollow structure without silicone filling leads to a reduction of 75$\%$ of the weight, resulting in an overall sensor weight of 420 g.
Due to the large and curved profile of the sensing surface, exploiting illumination sources spread at the contours of the silicone - like in previous approaches that feature a hollow structure \cite{tactile_bed} - would not properly allow the light to propagate all over the soft material. Thus, we place the light source outside the sensing area and distant from it to illuminate the whole surface homogeneously. Since regular LEDs would cause a large amount of reflections on the transparent rigid plate, we propose a novel solution based on the combination of UV LEDs, used as a light source, and a color filter applied on the camera. This solution makes the fluorescent green markers shine without significant reflections, so that they can be properly detected and tracked by the optical unit. Further, this solution does not lead to a substantial increase in the weight of the sensor, as it only requires two small UV LEDs strips and the filter. In addition, the outer faces of the rigid frame are covered to shield the sensor from external light (Fig. \ref{fig:design}C).

\vspace{-0.2cm}
\subsection{Fabrication}

The rigid frame of the sensor is made of 3 mm MDF panels (components n. 2 and 3 in Fig. \ref{fig:design}D) that fully encloses camera and illumination, and supports the sensing area. The transparent rigid plate is a stiff acrylic substrate (3 mm plexiglass) that is manually bent to a curved shape after heating it up (inner side of component 1 in Fig \ref{fig:design}D). The curvature is defined by a radius of 22 cm and an opening angle of 86$^rcirc$. Both the rigid MDF panels and the acrylic substrate are laser cut.

The curved sensing area (length equal to 32, width equal to 4 cm) is bonded to the transparent acrylic substrate, and made of a soft silicone layer (outer side of component n. 1 in Fig. \ref{fig:design}D). This layer (marker layer) of transparent silicone (Smooth-On Ecoflex Gel with Shore Hardness 000-35) hosts densely (and randomly) distributed markers (Cospheric Fluorescent Green Polyethylene Microspheres 1.02g/cc - 425um to 600um).
Using a 3D-printed mould, a mixture of markers and silicone is poured over the acrylic substrate.
\begin{figure*}[t]
    \centering
    \includegraphics[width=\linewidth]{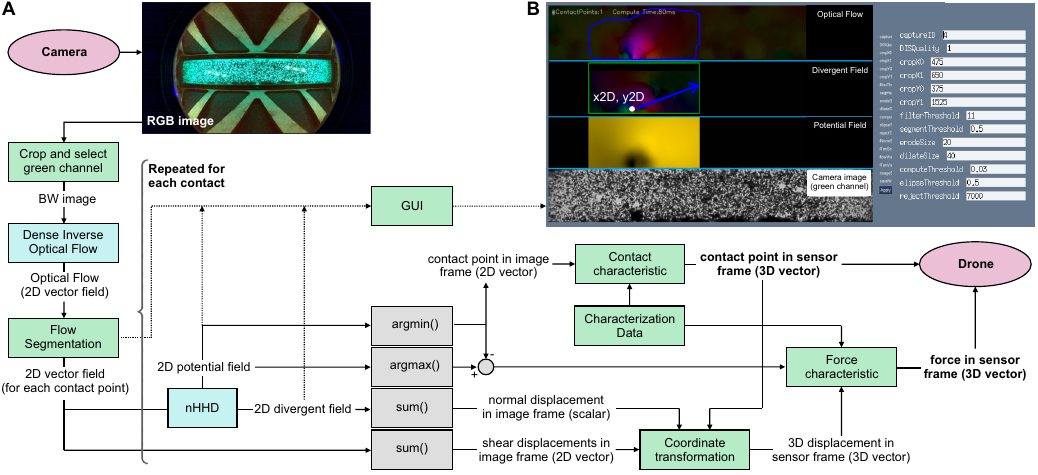}
    \caption{Flow-diagram for force and contact estimation. (A) Software pipeline developed for our optical tactile sensor: libraries shown in blue, mathematical functions in gray, contribution in green. (B) Graphical user interface: on the left, raw and divergent optical flow, potential field, and cropped camera image are displayed; on the right, the parameters used for cropping and thresholding can be changed to see the effect on the sensor performances.}
    \label{fig:programming_pipeline}
    \vspace{-0.2cm}
\end{figure*}
An external layer (coating layer) of opaque, black silicone (Ecoflex 00-30 with Shore Hardness 00-30 mixed with Elastosil Color Paste FL Black RAL 9011 pigments) is coated on the marker layer to block external light from interfering with the sensor. The markers are embedded in the marker layer and move with the deformation of the gel whenever a force is applied to its outer part. The motion of the markers is directly related to this force, which can therefore be reconstructed by tracking this motion. Avoiding a specific marker layout generally facilitates the development to larger surfaces of arbitrary shapes.

Two UV LEDs strips (AB-FC012UA-19700-XA2 LED strip, 7 cm long) are placed on the two internal sides of the structure (component n. 4 in Fig. \ref{fig:design}D). The color filter (SCHOTT GG-475, 25mm diameter, 3mm Thick, Colored Glass Longpass Filter from Edmunds Optics) is placed on the lens of the camera via a lightweight 3D-printed bracket (component n. 6 in Fig. \ref{fig:design}D).
The camera (component n. 5 in Fig. \ref{fig:design}D) is a ELP-SUSB1080P01-L180 (lens 180$^\circ$ FOV), which supports USB3.0 and uncompressed video at YUY2@50Hz. The distance between the camera and the sensing area is 120 mm, which ensures that the markers are not in the distorted area of the acquired images. The information at all pixels is processed, which allows to actually exploits the full resolution (1920$\times$1080) of the camera.
The 3D-printed connectors to attach the sensor to the drone (components n. 7 and 8 in Fig. \ref{fig:design}D) are made rigid enough to avoid bending, and leave enough room for the camera cable to come out of the structure. 
All the parts are rigidly assembled through connectors, 3D printed in ABS, and M2 screws. 
Finally, the whole sensor is covered with black paper and black electric tape to eliminate remaining external light interference, which could potentially influence the measurements. 
The final weight of the sensor is equal to 420 g. The main components and the assembled version are depicted in Fig. \ref{fig:design}.


\section{Multi-contact Sensing Methodology}
\label{sec:software}

An internal camera tracks spherical markers embedded in the silicone, which deforms whenever it enters in contact with the environment. We introduce a method to fully retrieve rich information about multiple contacts from images of the deformed sensing area. This is achieved by segmenting the optical flow obtained from the images, to consider several portions of the sensing area independently, and by exploiting the natural Helmholtz Hodge Decomposition (nHHD) \cite{Bhatia2014} to compute real-world position (mm) and 3D force (N) at multiple contact points.

A general description of the software pipeline is presented below and illustrated in Fig. \ref{fig:programming_pipeline}A:

\begin{itemize}
    \item a camera image is acquired with full resolution (1920$\times$1080) and at 48 fps;
    \item the image is therefore cropped and turned to a black and white image by selecting the green channel;
    \item the first cropped image is stored as a reference and new ones continue to get acquired; 
    \item the optical flow is computed from the images and used to calculate the relative movement of the markers from the first, reference image (where the markers are still and no force is applied) to the current one;
    \item the optical flow is therefore segmented into patches to isolate different contact points; 
    \item each patch is analyzed to estimate the center of the contact point location and the raw displacements applied (one normal and two tangential/shear components) by using the natural Helmholtz Hodge Decomposition (nHHD);
    \item the raw displacements are then incorporated into a polynomial, obtained from the characterization data, and transformed to real-world contacts forces and locations.
\end{itemize}

The software pipeline is fully implemented in both Python, with a user-friendly GUI, and in ROS, for running online and in real-time onboard. A filter can be enabled and included at the end of the software pipeline to smooth the variables of interest. The steps reported above are described in details in the following sections.

\vspace{-0.2cm}
\subsection{Optical Flow Processing and Segmentation}
\label{subsec:thresholds}

The optical flow is calculated using the Dense Inverse Search Optical Flow \cite{dense_inverse_search} function provided by the \textit{openCV} library. For this processing, we used the "fast" speed preset and standard parameters. Some preliminary testing showed no benefits in changing the parameters and a thorough investigation into all of the parameters is beyond the scope of this work.

Thus, the optical flow is segmented with a thresholding algorithm. First, the magnitude of the flow is calculated. Then, this magnitude field is thresholded to obtain a binary pixel mask. This roughly selects all areas where relevant deformations happen and excludes the rest. This threshold is therefore useful to discard false readings with a small intensity, which can occur due to noise. 

To further reduce noise, the mask is eroded (removing pixels around the regions selected by the pixel mask). The size of the deformed region in the silicone depends on the force and shape of the object. So if the optical flow is only in a tiny region above the threshold this could mean that either: i) the contact is very weak, ii) the object is very sharp like a needle, or iii) the increase is caused by noise. Thus, eroding the mask removes all areas below a given size. To compensate for the erosion, the mask is dilated afterwards. This ensures that the erosion did not split a narrow contact area in two and that the segmentation is not too restrictive and still contains all the parts of a contact region. 

After the erosion and dilation, the contours of the mask are found. Internal contours are ignored, since there can be no contact inside another contact. For each of the contours, a bounding box is obtained (Fig. \ref{fig:programming_pipeline}B). Thus, the absolute value of the flow inside this box is summed up and bounding boxes with the sum below a threshold are discarded. Hence, this threshold is useful to avoid outliers that can occur both when there is no contact and when there are already contacts detected. Finally, the unmodified optical flow is cut into patches matching the bounding boxes and each of the subflows is analyzed separately. 

Note that, on the one hand, lower thresholds can guarantee a lower minimum detectable force from the sensor readings. Having higher thresholds, on the other hand, can be helpful to reject wrong measurements, ensuring robustness to false contacts, which is desired during flight due to vibrations onboard. In Section \ref{sec:performance} we have reported the sensor performance for three options of segmentation thresholds, that we defined low, mid and high. For usage onboard, we decided to choose the last option for reducing outliers and noise, since the sensor is directly mounted on the frame of the drone.

\vspace{-0.2cm}
\subsection{Estimation of Raw Displacements}

For every subflow, we compute intermediate raw (scalar) quantities, denoted here as `raw displacements'. We define three components for such raw displacements, namely, we denote $D_{s1}, D_{s2}$ as tangential (or shear) components, and $D_n$ as the normal component. The tangential/shear components of the raw displacements are calculated by summing up, per axis, all the optical flow vectors within each subflow. To compute the normal component, instead, each subflow is processed with the natural Helmholtz Hodge Decomposition (nHHD) presented in \cite{Bhatia2014} (more details below), and the diverging vector field is extracted. This vector field is directly related to the normal force, as described in \cite{Zhang2019} for a single contact case. Therefore, we sum the absolute value of the divergence field vectors for each subflow to compute $D_n$. The next sections describe how the three components of raw displacements $(D_n, D_{s1}, D_{s2})$  are then transformed into our desired output quantities (Fig. \ref{fig:programming_pipeline}).

\textbf{Helmholtz Hodge Decomposition:} The Helmholtz Hodge Decomposition decomposes a vector field into a sum of three special vector fields: a divergence-free field, a curl free/potential field and a harmonic/uniaxial field. It is often used in analysing the motion of objects or fluids \cite{Guo2005}. In general, this decomposition requires a set of boundary conditions to obtain a unique solution. However, in \cite{Bhatia2014} the authors use a decomposition between external and internal flow to obtain a unique solution purely based on the data of the flow without any boundary conditions. They denote it as \textit{natural} Helmholtz Hodge Decomposition (nHHD) and provide a Python and C++ library online (\texttt{\url{https://github.com/bhatiaharsh/naturalHHD}}) to compute it seamlessly.

\vspace{-0.2cm}
\subsection{Estimation of Contact Points}

The potential field obtained from the nHHD is used to calculate the center of the contact points $(x_{2D}, y_{2D})$ (Fig. \ref{fig:programming_pipeline}B).
For a normal force, one would expect the displacement of the markers to be symmetrical with respect to the contact point. This means that the natural way to define the contact point is the center of the divergent vector field - the ``center'' is the local minimum of its potential. This means the contact point can be obtained by a simple \texttt{argmin()} call.

\vspace{-0.2cm}
\subsection{Calculation of Real Contact Location and Force}

The coordinates of the contact point obtained in the last section are in the frame of the camera image, with unit in pixel. To provide Cartesian values to the drone, a direct map from pixel to millimeters, as well as a transformation in the reference frame of the sensor, are required - this information can then be rotated into the drone's body frame or the world frame depending on the operating scenario. 
We define this map by expressing the x and y components of the real 3D contact location ($\hat{x}_{3D}, \hat{y}_{3D}$, measured in mm) as functions of $x_{2D}$ and $y_{2D}$ (in pixels), respectively. The z-coordinate $\hat{z}_{3D}$ is calculated using the radius of the sensor circular surface ($R =$ 0.22 m) and the distance ($d =$ 0.06 m) between the center of this circle and the sensor frame. This is summarized in the following:
\begin{equation}
\begin{aligned}
\begin{cases}
    \hat{x}_{3D} & = g_1 (x_{2D}) \\
    \hat{y}_{3D} & = g_2 (y_{2D}) \\
    \hat{z}_{3D} & = \sqrt{R^2 - \hat{x}_{3D}^2} - d
    \end{cases}
\end{aligned}
\label{eq:x3D}
\end{equation}
where the functions $g_1, g_2$, are computed experimentally by characterizing the sensor (see section \ref{sec:characterization}). 

Once the real contact location has been estimated, the raw displacements $(D_{s1}, D_{s2}, D_n)$ are transformed from the contact point to the sensor frame using the corresponding rotation matrix. Finally, we define the 3D force ($\hat{F_x}, \hat{F_y}, \hat{F_z}$) in the sensor frame (measured in N) as a function of such rotated displacements $(D_x, D_y, D_z)$, the contact location $(x_{2D}, y_{2D})$, and the difference between the maximum and the minimum potential field $d_{PF}$. The latter gives us information proportional to the size of the deformed area; thus, with this addition the fitting can provide an accurate estimation in case the drone interacts with obstacles of different size. This results in the following map:
\begin{equation}
\begin{aligned}
\begin{cases}
    \hat{F_x} & = g_x (D_x, D_y, D_z, x_{2D}, y_{2D}, d_{PF}) \\ 
    \hat{F_y} & = g_y (D_x, D_y, D_z, x_{2D}, y_{2D}, d_{PF}) \\
    \hat{F_z} & = g_z (D_x, D_y, D_z, x_{2D}, y_{2D}, d_{PF})
\end{cases}
\end{aligned}
\label{eq:F3D}
\end{equation}
where the functions $g_x, g_y, g_z$ are also computed through sensor characterization, as reported in Sec. \ref{sec:characterization}.

\vspace{-0.2cm}
\subsection{Graphical User Interface}

We developed a graphical interface to study the effect of different parameters on our pipeline. This is useful to adapt the minimum detectable force, the robustness to outlier, or the detection performance, depending on the specific needs. Our simple GUI is shown in Fig. \ref{fig:programming_pipeline}B. In the first row, the raw optical flow is displayed. The blue lines are the contours obtained from flow segmentation. The direction is signified by the color and the magnitude by the intensity. The second row shows the divergent component of the optical flow displayed in a similar way and surrounded by a bounding box. The contact location is shown by the white dot, whereas the direction and magnitude of the raw displacements are shown with the blue arrow originating from the point. The third row shows the corresponding potential field, which is interesting as it make easy to visualize the shape of the object pressing on the sensing area. The fourth row shows the camera image after cropping and converting to black and white. On the right, there is a user-friendly interface to tune the different thresholds. 


\section{Characterization}
\label{sec:characterization}

\begin{figure}[t]
    \centering
    \includegraphics[width=\linewidth]{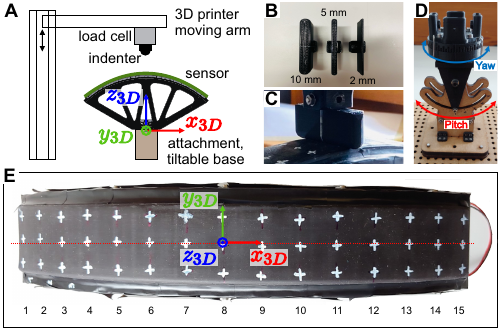}
    \caption{The setup used for characterization of the optical tactile sensor. (A) The sensor is attached to a base plate, placed on the table; the base can be tilted to rotate the sensor in multiple orientations. A load cell is connected to the vertical axis of a 3D printer, at the top, and to the indenter, at the bottom. A custom software framework allows to autonomously press the indenters on the sensing area. (B) Closeup of the cylindrical indenter with diameter equal to 10 mm pressing on the sensor. The indenter covers the whole sensor width since it is centered with respect to it. (C) Indenters used for the characterization; diameter of the cylinders is equal to 10 mm, 5 mm, and 2mm. (D) 2-axis tilting base used to hold the sensor in different orientations. Pitch and yaw rotations are reported. (E) The acquisitions are performed on the sensing area along the x-axis, in the center of the sensor's width, i.e. measurements on the numbered rows and centered on the dashed line.}
    \label{fig:characterization}
    \vspace{-0.2cm}
\end{figure}

The objective of the characterization is to find the functions described in Eqs. \ref{eq:x3D} and \ref{eq:F3D}. In particular, we restrict such functions to polynomial, reducing the task to retrieving their order and coefficients. We use these polynomials to map quantities estimated from the optical flow (raw displacements, contact location in pixels, potential field) to real-world values (contact location in mm and forces in N) . To this end, we apply a set of ground-truth forces (3-axis) with an indenter across the sensor surface and from multiple directions, and record the data of interest from the sensor (Fig. \ref{fig:characterization}A). 
As a result, we obtain a small yet sufficiently representative dataset that we use in a supervised learning fashion to fit such polynomials. A remarkable advantage of such a data-driven approach based on optical flow features is that it is suitable to achieve high accuracy while ensuring real-time inference and low modeling and data collections efforts. We choose this approach in contrast to i) FEM-based characterization \cite{taclink1, sferrazza2019ground}, which would require to fully model the flexible behavior of the sensing area, and ii) end-to-end learning-based approaches, which instead necessitate a huge volume of data to train a model that fits the data - while simulators can facilitate the collection of large datasets, it is nontrivial to obtain a very accurate model and to keep the \textit{sim2real} gap small \cite{Luu2023, sferrazza2022sim}. 

\vspace{-0.2cm}
\subsection{Data Collection}

Drones often encounter obstacles or interact with structures having cylindrical-shapes, such as tubes and pipes during industrial inspection and maintenance, as well as twigs and branches for monitoring and sampling in natural environments. Therefore, we propose to characterize the sensor with cylindrical indenters with three different diameters of the base (2 mm, 5 mm, 10 mm, Fig. \ref{fig:characterization}B). 
We restrict the cylindrical indenter to fully press on the silicone from one edge to the other along the width dimension (Fig. \ref{fig:characterization}C).
To apply forces from different directions we design a 2-axis rotating base to yaw and pitch the sensor into different angular positions (Fig. \ref{fig:characterization}D). We discard the roll rotation since it would cause the indenter to lose contact with the silicone and touch the rigid edges of the sensor's frame. After manually placing the base plate below the setup, the indenter is pressed down vertically on the sensor and an automatic software applies a set of forces ([0.5, 1, 1.5, 2, 2.5, 3] N in magnitude of the 3D force vector) - the printer stops at each of these thresholds to get a reliable measure. The measurements from a load cell attached to the setup, the information from the sensor camera, and the ground truth contact location (known since the indenter presses on the points indicated in Fig. \ref{fig:characterization}E) are saved in a synchronised way; these readings constitutes a single test. 

The datapoints are recorded along the x-axis of the sensor and centered with respect to the sensor's width - i.e. the indenter is placed on the numbered rows and centered on the dotted line (Fig. \ref{fig:characterization}E). For this reason the y component of the contact location becomes irrelevant for our fitting strategy, i.e., we set $\hat{y}_{3D}=g_2(y_{2D})=0$ over the entire data collection.
15 measuring points are marked, each spaced 20 mm apart from each other, but only 13 are used for the dataset (10 for training and 3 for validation and testing) - the two edges (first and last measuring points) are discarded as the silicone is in contact with the insulated tape applied to fully shield the inside from light so its properties are altered. 5 tests are repeated over each measuring point and for each of the three indenters. 
Despite the symmetrical shape of the sensor, the silicone may have a different behavior on the two sides of the sensor (one side is from line 2 to 7, the other one from line 9 to 14) due to fabrication asymmetries. Thus, we cover different combination of angles, symmetric in both sides of the sensors and values of angles, in order to be sure that the characterization dataset compensates for these asymmetries. The combination of sensor inclinations used are: 0$^\circ$ yaw and 0$^\circ$ pitch, 0$^\circ$ yaw and 15$^\circ$ pitch, 0$^\circ$ yaw and -15$^\circ$ pitch, 15$^\circ$ yaw and 0$^\circ$ pitch, -15$^\circ$ yaw and 0$^\circ$ pitch. This allows us to cover different values of forces on all the axis. We do not consider higher angles as they would lead to an incorrect interaction between the indenter and the sensing area (e.g. the indenter would touch sideways). 

The total amount of collected tests is equal to 855, later split into training (rows 2, 3, 4, 6, 7, 9, 10, 12, 13, 14 in Fig. \ref{fig:characterization}E) and validation (rows 5, 8, 11). Additional 210 test are acquired for testing (rows 5, 8, 11) over the sensing area.

\vspace{-0.2cm}
\subsection{Fitting Methodology}

Here, we describe the procedure to map the quantities extracted from the optical flow to the x-component of the contact location ($\hat{x}_{3D}$) in mm and 3D contact force vector ($\hat{F_x}, \hat{F_y}, \hat{F_z}$) in N, both defined in the sensor frame. As mentioned above, the y component of the real contact location ($\hat{y}_{3D}$) is set to zero, since we only consider contacts covering the entire width of the sensor. The z component ($\hat{z}_{3D}$) can instead be computed analytically from Eq. \ref{eq:x3D} once $\hat{x}_{3D}$ is known.

For each test we extract the recorded data: the load cell provides a 3D vector of ground truth force $F_x, F_y, F_z$, which is opportunely rotated in the sensor frame, knowing the yaw and pitch angle during the specific tests; the processed information of the sensor, instead, returns a 3D vector of raw displacement, which is also rotated in the sensor frame to get ($D_x, D_y, D_z$), the location of the contact $x_{2D}$ in pixels, and the difference between the max and min value of potential field $d_{PF}$. 

We divide the entire dataset in three sets depending on the measuring points (as described above), obtaining a training set, a validation set, and a testing set. Therefore, we perform a polynomial fitting to find the coefficients of the polynomials. In detail, for the function $g_1$ in Eq. \ref{eq:x3D}, we set a maximum order and perform the fitting for every order from 1 to the maximum. We select the optimal order as the one that minimizes the validation error during training. As a metric for the error we use the Mean Absolute Error (MAE). The same procedure is performed to find the functions $g_x, g_y, g_z$ of Eq. \ref{eq:F3D}; in this case we use different orders for subsets of variables. Summarizing, at the end of this step, different optimal orders are found for the raw displacements, the contact location in pixel, and the potential field.
The whole processing and data analysis procedure has been executed in Matlab. The polynomials obtained from the characterization, both in terms of coefficients and orders, have been integrated in the software pipeline to compute contact locations and forces in real-time.


\section{Sensor Performance}
\label{sec:performance}

The performance of the sensor is evaluated over the testing set (as aforementioned, rows 5, 8, and 11 in Fig. \ref{fig:characterization}E).

First, we assess the performance of the sensor for different segmentation thresholds used in our pipeline (Sec. \ref{subsec:thresholds}). We define three values as low, mid and high. As previously discussed, lower thresholds improve the minimum detectable force of the sensor, whereas higher thresholds can guarantee more robust measurements against noise and outliers during flight, therefore the latter is preferable. Table \ref{tab_thresholds} highlights that the minimum detectable force $F_{min}$ increases for higher values of thresholds, as expected. Moreover, the analysis shows that the prediction errors for the contact location and forces, expressed with the Mean Absolute Error (MAE), are comparable for the different thresholds - changes are in the order of 0.1 mm and 0.001 N.

\begin{table}[h]
\begin{center}
\caption{ERRORS ON TESTING SET FOR DIFFERENT THRESHOLDS.}
\label{tab_thresholds}
\begin{tabular}{| c | c | c | c |}
\hline
\multirow{2}{*}{ \textbf{Variable}} & \textbf{Low} & \textbf{Mid} & \textbf{High} \\
 & \textbf{Thresholds}  & \textbf{ Thresholds} & \textbf{ Thresholds} \\
\hline \hline
$F_{min}$ [N] & (0, 0.5) & [0.5, 1) & [1, 1.5) \\
\hline
MAE $\hat{x}_{3D}$ [mm] & 1.4 & 1.5 & 1.5 \\
\hline
MAE $\hat{F_x}$ [N] & 0.071 & 0.074 & 0.070\\
\hline
MAE $\hat{F_y}$ [N] & 0.024 & 0.024 &  0.026 \\ 
\hline
MAE $\hat{F_z}$ [N] & 0.169 & 0.166 & 0.172 \\
\hline 
\end{tabular}
\end{center}
\end{table}

This result suggests that using higher thresholds allows to maintain the same performance compared with lower ones. However, the drone will be able to detect only forces above a certain value. It is worth to notice that our selection of ``high'' thresholds permits to detects forces as small as 1 N, with the maximum applied force in our dataset equal to 3 N, making sure that the dataset remains sufficiently representative.

Second, upon selecting the set of thresholds as discussed above, we report the prediction errors for the different points of the testing set (Tab. \ref{tab_error}).  
The analysis highlights that the errors in terms of contact locations are the lowest for point 8, which is in the center along the sensor length. The errors for longitudinal and lateral forces vary within the same order of magnitude, with the higher value of error being very small (0.078 N for point 11). The errors for the vertical force are one order of magnitude bigger than the other two components, yet the same components are in the same order of magnitude  among the contact points (the highest error is equal to 0.23 N and associated with point 11). It is worth to mention that the range of forces in the three components is different, spanning [-1, 1] N for x-y components and [-3, 0] N for the z component.

\begin{table}[h]
\begin{center}
\caption{ERRORS ON TESTING SET FOR THE SELECTED THRESHOLDS.}
\label{tab_error}
\begin{tabular}{| c | c | c | c | c |}
\hline
\textbf{Variable} & \textbf{Total} & \textbf{Point 5} & \textbf{Point 8} & \textbf{Point 11} \\
\hline \hline
MAE $\hat{x}_{3D}$ [mm] & 1.5 & 1.8 & 1.0 & 1.6 \\
\hline
MAE $\hat{F_x}$ [N] & 0.070 & 0.059 & 0.074 & 0.078 \\
\hline
MAE $\hat{F_y}$ [N] & 0.026 & 0.026 & 0.024 & 0.029 \\ 
\hline
MAE $\hat{F_z}$ [N] & 0.172 & 0.103 & 0.196 & 0.230 \\
\hline 
\end{tabular}
\end{center}
\end{table}

Despite the errors being overall very similar across contact points, considering each single variable, we ascribe the small varying performance to the pattern variations introduced by the fact that the markers are randomly distributed into the silicone - in some area their distribution might be easier to be tracked by the algorithm upon deformation. 

\begin{figure}[t]
    \centering
    \includegraphics[width=\linewidth]{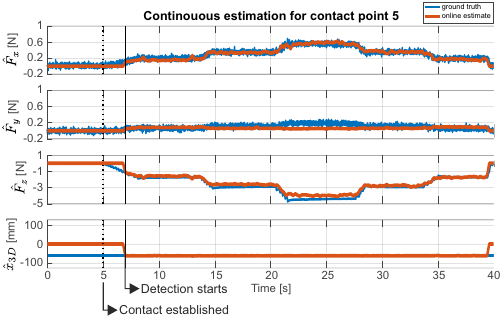}
    \caption{Estimation of a single contact continuously applied at position 5. Ground truth from the load cell, force and contact location estimated (and filtered) online with our software pipeline.}
    \label{fig:online_estimation}
    \vspace{-0.2cm}
\end{figure}

Third, we validate the performance of the sensor by applying  a continuous, time-varying force on the sensing area and estimated the contact location and force online and in real-time. As an example, Fig. \ref{fig:online_estimation} depicts the time evolution of $\hat{x}_{3D}, \hat{F_x}, \hat{F_y}, \hat{F_z}$ compared to the ground truth, for a test executed with zero pitch and zero yaw angles, on contact point number 5 over the sensing area (-60 mm from the center of the sensor frame). The plots validates the accuracy of the full pipeline in the estimation of the quantities of interest over time. Furthermore, the detection starts around a time of 7 seconds in Fig. \ref{fig:online_estimation}, when a 3D force (in magnitude) higher than 1 N is applied on the sensing area. This is in accordance with the choice of the thresholds previously described.

\begin{figure}[t]
    \centering
    \includegraphics[width=\linewidth]{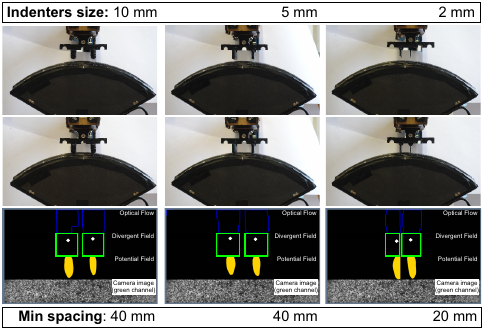}
    \caption{Minimum spacing between two contacts to be successfully distinguished. Experimental setup with multiple indenters (before and during contacts) and GUI showing the two contacts separated from each other. The minimum spacing varies depending of the size of the indenter.}
    \label{fig:min_spacing}
    \vspace{-0.2cm}
\end{figure}

Finally, we conducted quantitative tests with multiple indenters attached to the characterization setup, to understand the minimum spacing between two contacts that allows to properly separate between them (Fig. \ref{fig:min_spacing}). We tested for indenters having diameter equal to 2 mm, 5 mm, and 10 mm - the same used for the characterization. The minimum spacing between two contacts changes depending on the size of the indenters: for the 5mm and 10 mm indenters the minimum spacing is equal to 40 mm, whereas with the smallest indenter (2 mm), the minimum spacing is equal to 20 mm. The latter means that the algorithm can generally detect two contacts if they are spaced at least 40 mm apart, which is remarkable with respect to previous works. Figure \ref{fig:min_spacing} summarizes these results by showing the setup with multiple indenters (before and during contacts), the two contacts well distinguished on the GUI, and the values of minimum spacing for different indenter sizes.

\section{Experimental Validation}
\label{sec:experiments}

\subsection{Drone architecture}
\label{subsec:drone}

Our aerial robot is a quadrotor that consists of a 6-inches carbon-fiber frame, brushless motors DYS Thor 2408- 2200KV with 6-inch 2-blades propellers (each providing about 8 N lift force), a BetaFlight Flight Controller (FC) for attitude control, and a Khadas VIM3 companion computer for high-level control, communication, and additional on-board computations. The system is powered with a 3-cell 2500 mAh LiPo battery and has a total mass (including the sensor) of around 1.2 kg. The optical tactile sensor is rigidly mounted to the drone's frame with a 3D printed connector. To fully work onboard, it is only required to connect the camera unit to a USB port of the onboard computer, and to power the UV LEDs.

\vspace{-0.2cm}
\subsection{Demonstration}
\label{subsec:demos}

\begin{figure}[t]
    \centering
    \includegraphics[width=\linewidth]{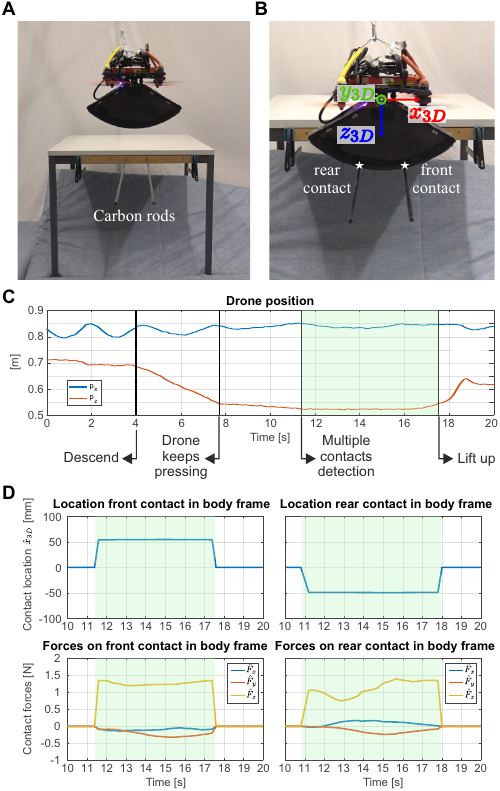}
    \caption{Detection of multiple contacts and estimation of their location and forces. (A) Experimental setup with two perches (carbon rods) attached to a rigid structure (table). (B) Example of an experiment where the drone descends on the perches and keeps contact with them. The sensor detects the two contacts during the interaction (green area). Contact points highlighted with a star symbol; body coordinate frame reported in the center-of-mass of the drone. (C) Drone's global position in x and z directions. The different phases of the demo are highlighted. (D) Close-up plots to show the locations and 3D forces of the two contacts, measured onboard in the body frame.}
    \label{fig:demo_multicontact}
    \vspace{-0.2cm}
\end{figure}

Here, we demonstrate the versatility of the sensor and its use in multi-contact scenarios, by considering tasks where both 3D contacts and forces detection are relevant. The demos involve interaction with stiff and soft carbon rods, i.e. structures with a cylindrical shape, in accordance to the explanation provided in the Introduction and Characterization chapters. We use carbon rods (Fig. \ref{fig:demo_multicontact}A) with diameters in the range considered during the characterization (between 2 mm and 10 mm).

To start, we validate that the drone effectively integrates the sensor onboard to estimate multiple contacts location and force in real-time while touching two rods. 
Then, we further demonstrate two applications, showing how the drone can exploit the sensor readings. 
The first demo relates to the estimation of the compliance of two rods and re-alignment with the stiffer one. For instance, information on the compliance can be useful to identify a more stable perch where to rest, as drones can get a better support on stiffer rods, both in terms of stability and energy consumption, as demonstrated in \cite{aucone}.
The second demo involves the mapping of the global location of sparse rods that obstruct its path. This can be useful when navigating in cluttered areas, as drones are able to distinguish safe and empty spaces from cluttered areas.
The demos have been implemented in ROS 
and then transferred to the real drone, on the onboard computer. \\

\textbf{Detection of multiple interactions}: For this demo (Movie S1), we command the drone to autonomously descend on two rods (same diameter, Fig. \ref{fig:demo_multicontact}A) and keep contact with them (Fig. \ref{fig:demo_multicontact}B). Figure \ref{fig:demo_multicontact}C depicts the global position of the drone in x and z direction. It is possible to see that the drone is commanded to descend vertically, as the x does not change whereas the z decreases. Upon contact with the rods, it keeps pressing down;  due to the thresholds used to reduce noise onboard, the drone has to press with a force higher than 1 N to properly detect both contacts. At this point, after the two forces are high enough so that the algorithm can detect them, the drone stops pressing and keeps that condition for a predefined amount of time, during which the location and forces of the two rods in the body frame are measured (green highlight) and the drone does not move (x and z positions are constant). The body frame is selected as coincident with the sensor frame. Figure \ref{fig:demo_multicontact}D shows a close up of the contacts detection, depicting the x-location of the two contacts - which are mirrored in values as the drone was centrally aligned above the two perches before descending - and the three values of force in x, y, and z direction. The main contribution comes from the z-component due to the vertical descend, but the sensor detects also a lateral contribution due to a small bending of the perches. \\

\begin{figure}[t]
    \centering
    \includegraphics[width=\linewidth]{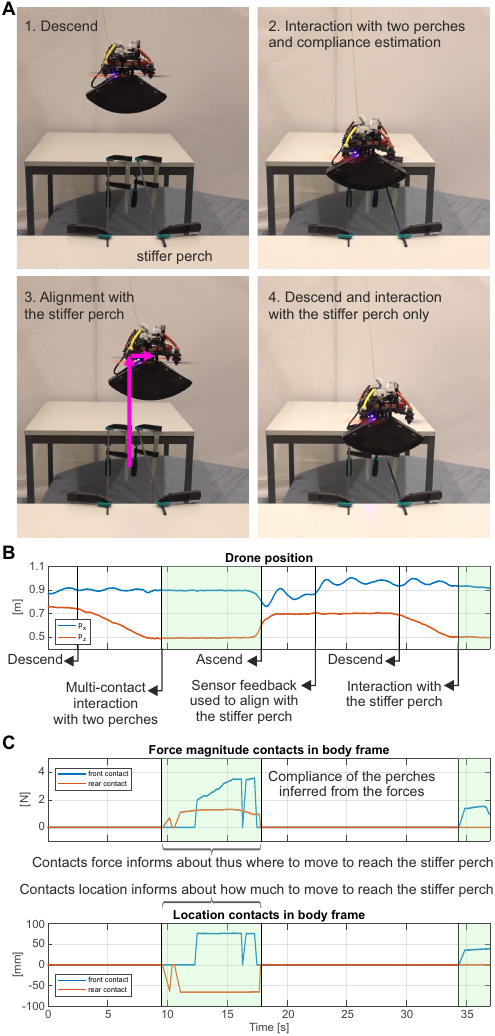}
    \caption{Compliance estimation and landing after re-alignment. (A) Snapshot of the drone during different phases of the demo. (B) Drone's global position in x and z directions. The different phases of the demo are highlighted. (C) Plots of the variables of interest over time, showing the location and force of the two contacts measured onboard in the body frame, during the first interaction, and location and force of the single contact with the stiffer perch after the re-alignment and landing.}
    \label{fig:compliance}
    \vspace{-0.2cm}
\end{figure}

\textbf{Compliance estimation, re-alignment, and landing}: The objective is to validate how drones can exploit the information from the sensor for decision making depending on the environment (Fig. \ref{fig:compliance} and Movie S2). To this end, we demonstrate that upon contact with two perches the feedback from the sensor allows the drone to distinguish which perch is stiffer and where it is located; thus, knowing the compliance of the perches allows the drone to decide in which direction to fly, whereas the information on the position of the stiff perch defines how much it has to move to be aligned above it for landing. In detail (Fig. \ref{fig:compliance}A), the drone is commanded to autonomously descend, starting from a position centrally aligned with the two perches that have different compliance. Thus, the drone maintains contact with them and measures which one is stiffer; we assume that the perches deform equally during the interaction, hence the one producing a higher force is the stiffer - the higher the force the lower the compliance. Then, the drone flies back up and autonomously re-aligns on top of the stiffer perch by using the information about the location of the stiffer contact. Finally, the drone autonomously lands on it and maintain contact. Figure \ref{fig:compliance}B highlights the aforementioned phases of the demo alongside the position of the drone, where the descends are associated with a decrease in the z-position, and the re-alignement with an increase in the x-position. The interaction is highlighted with green areas. Fig. \ref{fig:compliance}C reports the force magnitude and the location of the two contacts. The force magnitude shows a higher value on the front contact, which relates indeed to the stiffer perch. Regarding the contacts location, during the first descend the sensor detects both contacts, whereas after re-alignment it correctly measures only the single contact occurring on the stiffer perch.\\

\begin{figure}[t]
    \centering
    \includegraphics[width=\linewidth]{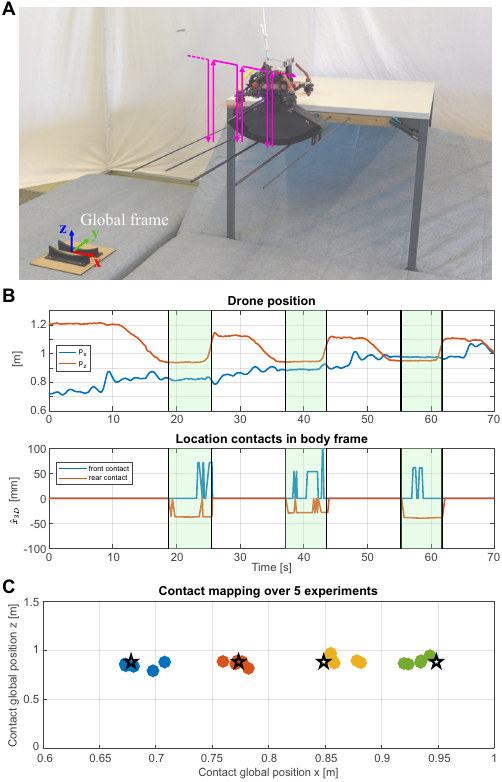}
    \caption{Contact mapping demo. (A) Snapshot of the drone touching and mapping multiple perches. Reference trajectory of the drone and global frame highlighted. (B) Position of the drone and estimated contacts location in body frame over time during the demo. (C) Location of multiple perches in the global frame, estimated onboard upon physical interaction, over 5 experiments.}
    \label{fig:mapping}
    \vspace{-0.2cm}
\end{figure}

\textbf{Mapping by touch}: We place multiple rods of different diameter in the scene, with the aim of showcasing the successful detection and mapping of such obstacles by direct physical interaction (Fig. \ref{fig:mapping} and Movie S3). The drone is commanded to descend, to keep contact for a predefined amount of time, then move up and forward for a predefined distance, and to repeat the operation autonomously. The objective is to simultaneously estimate the location of the obstacles in a global frame in order to map the environment. Figure \ref{fig:mapping}A captures the drone during the demo, with the reference trajectory to follow highlighted in pink. Thus, Fig. \ref{fig:mapping}B shows the global position of the drone and the locations of the detected contacts (in the body frame) over time. During the detection, the contact locations are transformed (and stored) in the global frame by using the position and orientation of the drone at that instant of time. Five experiments are conducted to assess the repeatability of the demonstration. For all the tests, the estimated contact locations in the global frame (in x and z directions) are reported and highlighted in different colors in Fig. \ref{fig:mapping}C, showing an estimation errors lower than 4 cm when compared to the ground truth.
During all the interactions, the drone keeps contacts with at least two perches. It is worth to notice that the sensor readings sometimes jump to zero during the detection (Fig. \ref{fig:mapping}C): this is not due to loss of contacts, but rather because the values of the interaction forces go slightly below the minimum detectable force threshold for some instants/samples. This did not happen in the previous demos as the drone was centered with respect to the mid point between the two rods, allowing to properly keep contact with them. For this demo, instead, the drone is randomly aligned above the rods before descending. In particular, due to this misalignment, the interaction with the last rod has not been consistent among the experiments, leading to an unstable interaction or an unsuccessful detection in some cases. This is why the setup presents 5 rods but only the first 4 have been used for the analysis. Nevertheless, this could be solved by implementing an external controller that ensures a stable physical interaction with a desired force, e.g. an admittance controller that exploits the sum of all the forces, but this is beyond the scope of the article.  

\section{Conclusion}
\label{sec:conclusion}

Our work aims to bring distributed tactile sensing technology to the domain of aerial robotics. We have developed a curved, large-scale sensor as big as an arc of a circular ring, consisting of a sensing area of 32 cm by 4 cm, a hollow structure, and a novel illumination system. Thanks to these features, the sensor can convey detailed information regarding contact location and force distribution from multiple interactions between the robot and its surroundings. Our software pipeline permits to simultaneously detect multiple contacts spaced as close as 2 cm, providing real-world quantities in the form of contact locations (mm) and force vectors (N) with an accuracy of 1.5 mm and 0.17 N, respectively. After integration onboard our custom drone, the usability and reliability of the sensor has been validated in real-time through multiple demonstrations. The experiments demonstrates how information from the sensor can enable drones' ability to perform a variety of different tasks, such as detecting and interacting with multiple obstacles, estimating their compliance, mapping them.

One future development of our work lies in the design of the sensor. The proposed mechanical design presents decreased capabilities to measure contacts on the edges of the sensing area. Increasing the width of the sensor or designing a surface with a curvature along the sensor's width could allow for detection of more contacts from different directions, as well as enlarge the range of forces that can be detected. Moreover, our current prototype limits the sensing capabilities to the bottom part of the drone. A spherical shape could potentially permit to fully enclose the drone and provide sensing capability all around its body, scaling the usage towards interaction with more obstacles sparsely distributed in the cluttered environment.

To perform dynamic tasks such as sliding along surfaces or traversal of obstacles \cite{traversal}, the high friction of soft silicone skin in direct contact with the environment could hamper the drone's capability. The addition of a low-friction coating or a thin polymer film on the silicon skin would mitigate this problem.

Another future direction could involve the usage of the potential field obtained from the nHHD to estimate the orientation and the size of the rods interacting with the sensing surface. Drones could exploit the first information to re-align their orientation with respect to very sparse obstacles, and the second information to enhance the estimation of the compliance. From a broader perspective, after multiple detections drones could incorporate these quantities into a perception framework to improve the overall map of the environment.

Finally, the successful integration and utilization of the sensor further inspire contributions to novel control strategies. Taking inspiration from humanoids, successfully exploiting multiple contacts could prove beneficial to various aerial physical interaction tasks, e.g. applying different forces in multiple interaction points, or moving objects along different paths for non-prehensile manipulation. We anticipate that the synergy between distributed tactile sensing and touch-based control strategies will play a crucial role in the use of drones in the real world, expanding their viability for aerial physical interaction and navigation in visually degraded environments.

\end{document}